# Leveraging Unlabeled Scans for NCCT Image Segmentation in Early Stroke Diagnosis: A Semi-Supervised GAN Approach

Maria Thoma
*Dept. of Computer Science and Biomedical Informatics*
*University of Thessaly*
Lamia, Greece
thmaria@uth.gr

Michalis A. Savelonas
*Dept. of Computer Science and Biomedical Informatics*
*University of Thessaly*
Lamia, Greece
msavelonas@uth.gr

Dimitris K. Iakovidis
*Dept. of Computer Science and Biomedical Informatics*
*University of Thessaly*
Lamia, Greece
diakovidis@uth.gr

***Abstract*— Ischemic stroke is a time-critical medical emergency where rapid diagnosis is essential for improving patient outcomes. Non-contrast computed tomography (NCCT) serves as the frontline imaging tool, yet it often fails to reveal the subtle ischemic changes present in the early, hyperacute phase. This limitation can delay crucial interventions. To address this diagnostic challenge, we introduce a semi-supervised segmentation method using generative adversarial networks (GANs) to accurately delineate early ischemic stroke regions. The proposed method employs an adversarial framework to effectively learn from a limited number of annotated NCCT scans, while simultaneously leveraging a larger pool of unlabeled scans. By employing Dice loss, cross-entropy loss, a feature matching loss and a self-training loss, the model learns to identify and delineate early infarcts, even when they are faint or their size is small. Experiments on the publicly available Acute Ischemic Stroke Dataset (AISD) demonstrate the potential of the proposed method to enhance diagnostic capabilities, reduce the burden of manual annotation, and support more efficient clinical decision-making in stroke care.**

## I. INTRODUCTION

Ischemic stroke, a condition resulting from the sudden blockage of an artery supplying blood to the brain, represents a potentially devastating medical event and stands as a leading cause of long-term disability and death globally [1]. The viability of brain tissue diminishes rapidly following an ischemic event, underscoring the critical importance of early and accurate diagnosis. Prompt therapeutic interventions, such as intravenous thrombolysis or endovascular thrombectomy, can significantly mitigate brain damage and dramatically improve patient outcomes, but their efficacy is critically dependent on the time elapsed since symptom onset. However, the clinical presentation of stroke can be ambiguous, overlapping with other neurological conditions, which makes diagnostic imaging an indispensable component of the emergency assessment workflow.

Non-contrast computed tomography (NCCT) has emerged as the cornerstone and first-line imaging modality for the initial evaluation of acute stroke [2]. Its principal advantages are its wide availability in most emergency departments, rapid acquisition time, and cost-effectiveness. The primary role of NCCT in the hyperacute setting is to reliably exclude intracranial hemorrhage, which is a contraindication for thrombolytic therapy. Despite its value, the interpretation of NCCT scans for early ischemic changes presents considerable challenges. In the first few hours - the crucial hyperacute phase - the signs of ischemia, such as cytotoxic oedema causing subtle hypoattenuation, can be extremely faint. This makes their detection highly dependent on the expertise of the interpreting radiologist and subject to inter-observer variability. This diagnostic uncertainty can lead to critical delays in treatment, motivating the development of computational methods to accelerate and objectify the analysis of NCCT scans.

Prior to the widespread adoption of deep learning, efforts to automate stroke lesion detection on NCCT scans relied on traditional image analysis. These methods included intensity thresholds, region-growing algorithms and more sophisticated model-based techniques like level-sets and atlas-based registration. They typically depended on handcrafted features designed to capture specific image properties like texture and intensity distributions. While effective in certain controlled scenarios, their performance was often limited in clinical practice. Their reliance on local intensity cues and limited semantic context made it difficult to reliably distinguish subtle, early ischemic changes from noise, artifacts, or natural anatomical variability, particularly in cases of small lesions or poor contrast.

Recent advances in machine learning have significantly enhanced the early identification and assessment of acute ischemic stroke using NCCT. Radiomics-based approaches have obtained promising results for the detection of ischemic changes that are imperceptible to radiologists, as is the case with the work of Wu et al. [3], who employed radiomics features from symmetric image patch pairs, sparse representation and multi-scale maximum a-posteriori probability (MAP) classification. The advent of deep learning, particularly convolutional neural networks (CNNs) [4], has led to significant advances in medical image segmentation. Architectures, such as U-Net [5] and variants have been widely applied to stroke lesion segmentation, demonstrating a powerful ability to learn hierarchical feature representations directly from image data. These models have shown considerable success in segmenting subacute stroke lesions, which are more pronounced on NCCT. Kuang et al. [6] introduced early infarct segmentation network (EIS-Net) for NCCT image segmentation and early infarcts (EI) boundary extraction, as well as for computing the Alberta stroke program early CT score (ASPECTS). EIS-Net comprises a 3D triplet convolutional neural network (T-CNN). The nnU-Net model has recently demonstrated promising performance in segmenting NCCT images of patients with acute ischemic stroke [7]. Another U-Net-based model has been recently employed by Ostmeier et al. [8]. Their model was trained on reference annotations of three experienced neuroradiologists using majority voting and random expert sampling training schemes. In this light, generative adversarial networks (GANs) [9] offer a compelling alternative, since they can generate anatomically plausible artificial ground truth masks that can be used in a semi-supervised context.

In this work, we propose a semi-supervised adversarial learning method for the segmentation of early ischemic stroke lesions in NCCT scans, for which pixel-level annotations are inherently scarce. The proposed method is designed to leverage both a small set of labeled NCCT scans and a larger pool of unlabeled scans drawn from the same data distribution by employing semi-supervised learning in the form of adversarial training. Highly-confident predictions on unlabeled data are progressively incorporated as pseudo-labels. The proposed method addresses the challenges of early stroke segmentation more effectively by employing:

- a hybrid Dice and cross-entropy loss to tackle the class imbalance of small ischemic lesions,
- a feature matching loss to encourage the generator to produce anatomically plausible segmentation maps,
- a self-training loss to reinforce learning from high-confidence pseudo-labeled samples.

This approach allows the model to learn robust and generalizable features, even with limited supervision.

The remainder of this manuscript is structured as follows: Section II presents the proposed semi-supervised adversarial learning framework in detail. Section III describes the experimental setup, the dataset used for evaluation, and discusses the quantitative and qualitative results obtained. Finally, Section IV summarizes the main conclusions of this work and suggests directions for future research.

## II. Semi-supervised Adversarial Learning

The proposed method leverages the semi-supervised GAN-based learning strategy of the s4GAN network [10]. Although that network was originally developed for natural image segmentation, in the context of this work it is repurposed for generating masks that represent stroke lesions in NCCT scans, during the early stroke stages. The network is trained using a small set of labeled samples with corresponding ground truth masks, supplemented by a large pool of unlabeled NCCT scans that are confidently pseudo-labeled with artificially generated ground truth masks.

The proposed method employs SCUNet++ [11], which originally demonstrated a robust performance in detecting and delineating pulmonary emboli (PE) in computed tomography pulmonary angiography (CTPA) scans. Given the challenges associated with identifying PE lesions, mainly associated with their small size, SCUNet++ is a suitable choice for NCCT scan segmentation during the very early stage, where delineating hyperacute stroke lesions is difficult, due to subtle changes in tissue attenuation.

The s4GAN model consists of two main components: the generator and the discriminator, described in Section II.A and II.B, respectively. The generator operates as a segmentation network and is trained by both labeled and pseudo-labeled NCCT scans, as described. The discriminator receives an input image paired with its ground truth mask, either real or artificial. Its output is a probability indicating whether the ground truth mask is real or artificial. These two components are trained in an adversarial manner, guiding the generator towards realistic and accurate artificial ground truth masks. Figure 1 provides an overview of the proposed method.

For network training, multiple loss functions are utilized: i) a hybrid loss function combining the cross-entropy and Dice losses is used to enhance pixel-wise accuracy and address the imbalance between stroke lesion and background, ii) a feature matching loss function that encourages the predicted artificial masks to exhibit feature-level characteristics that are similar to those of the real ground-truth masks, and iii) a self-training loss function, applied only to unlabeled data by leveraging the most confidently pseudo-labeled artificial ground truth masks.

### *A. Generator*

The generator results in artificial ground truth masks that either delineate stroke lesions on NCCT scan slices or are empty in cases of absent stroke lesions. During training, the generator leverages a small set of labeled samples along with a large pool of unlabeled data.

The generator minimizes a hybrid loss function that combines cross-entropy and Dice losses, using the available ground truth masks. This combination aids the generator in resulting in accurate segmentation masks, despite the class imbalance associated with the small size of stroke lesions. This imbalance occurs both in the slice level, where the lesion occupies only a small fraction of the image, as well as in the scan level, where a large fraction of scans is normal. The cross-entropy loss function can be defined as follows:

$$L_{\mathrm{CE}} = -\frac{1}{N}\sum_{i=1}^{N}\sum_{c=1}^{C} y_{i,c}\left(p_{i,c}\right) \quad (1)$$

where $N$ denotes the number of pixels in an image, $C$ is the total number of classes, $y_{i,c} \in \{0,1\}$ represents the ground truth label for pixel $i$ and class $c$, and $p_{i,c} \in [0,1]$ is the probability of pixel $i$ being assigned to class $c$.

The cross-entropy loss function performs best when the class distribution is balanced. Aiming to mitigate the effects of class imbalance, we incorporate the Dice loss function. The Dice loss function can be defined as follows:

$$L_{\mathrm{Dice}} = 1 - DSC \quad (2)$$

where DSC refers to the dice coefficient provided by:

$$DSC = \frac{2\sum_{i=1}^{N} p_i g_i}{\sum_{i=1}^{N} p_i^2 + \sum_{i=1}^{N} g_i^2} \quad (3)$$

where $N$ is the number of pixels, $p_i$ is the probability that the $i$-th pixel belongs to the target class, and $g_i$ is the ground truth label of the $i$-th pixel.

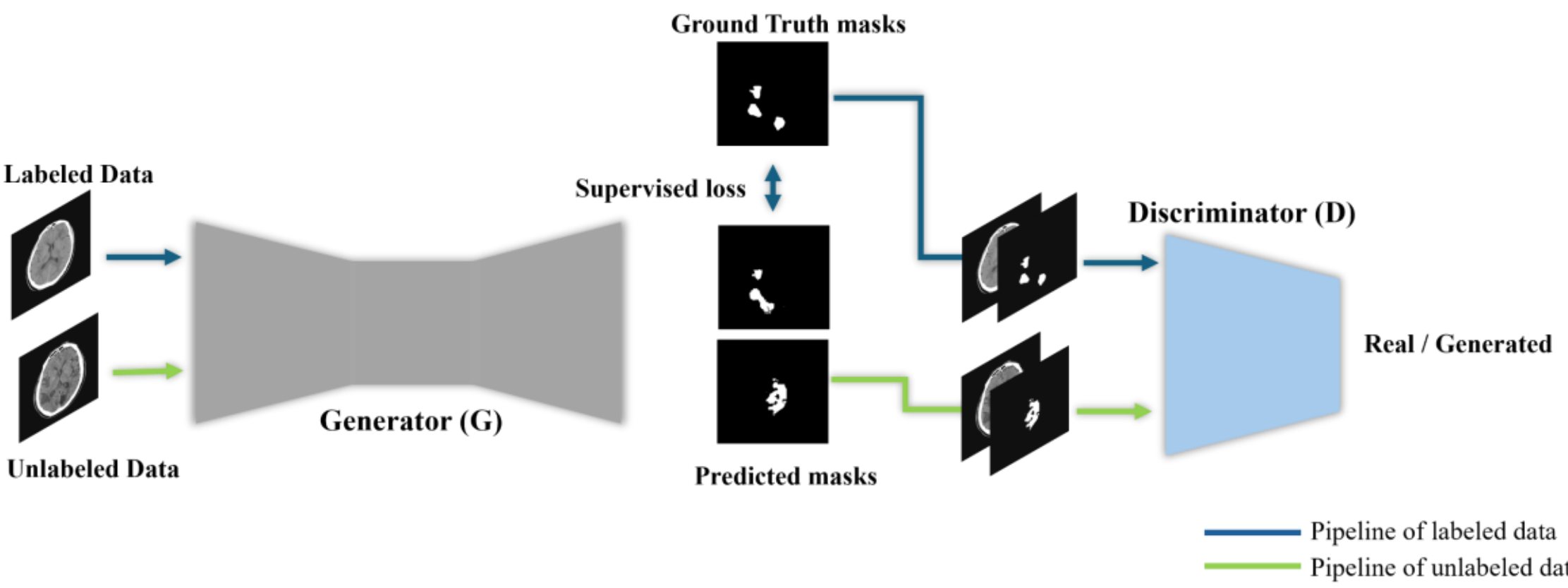

Fig. 1. Overview of the proposed S4GAN-IS method.

Aiming to leverage information from the large pool of available unlabeled NCCT scans, we also incorporate feature matching loss and self-training loss. The feature matching loss, as evident in Eq. (4), is formulated to minimize the discrepancy between intermediate feature representations of the real ground truth masks $y_L$ and the artificial ground truth masks $S(x_U)$. The feature matching loss function can be defined as:

$$L_{FM} = \left\| E_{(x_L,y_L)\sim D_L}[\Phi_k^D(y_L \oplus x_L)] - E_{x_U \sim D_U}\left[\Phi_k^D(S(x_U) \oplus x_U)\right] \right\| \tag{4}$$

where $\Phi_k^D(\cdot)$ refers to the feature maps extracted from the intermediate $k$-th layer of the discriminator $D$ and $\oplus$ denotes concatenation along the channel dimension.

The self-training loss function is applied only to unlabeled NCCT scans. The artificial ground truth masks $(y^*)$ provided by the generator are evaluated by the discriminator, which aims to distinguish between real and artificial ground truth masks, as illustrated in Fig.1. The self-training loss function can be defined as follows:

$$L_{ST} = \begin{cases} -\sum_{h,w,c} y^* \log \mathrm{S}(x_U) & , \text{if } D\big(\mathcal{S}(x_U)\big) \geq \tau \\ 0 & , \text{otherwise} \end{cases} \tag{5}$$

The discriminator output $D(\mathcal{S}(x_U))$ is a probability indicating how likely the input sample is labelled with a real ground truth mask. Pseudo-labels that exceed a threshold $\tau$ are considered to have "fooled" the discriminator and are therefore used in the computation of the self-training loss function.

The total loss function $Ls$ is computed as a weighted sum of the 4 previously defined loss functions and is formally expressed as follows:

$$L_s = w_{CE} L_{CE} + w_{Dice} L_{Dice} + w_{FM} L_{FM} + w_{ST} L_{st} \tag{6}$$

where $w_{CE}$, $w_{Dice}$, $w_{FM}$, $w_{ST}$ are the weights of $L_{CE}$, $L_{Dice}$, $L_{FM}$ and $L_{ST}$ respectively, adjusting the contribution of each loss term.

## B. Discriminator

The purpose of the discriminator is to distinguish between real and artificial ground truth masks, thereby guiding the generator to produce more accurate artificial ground truth masks. It is trained using as input the real data, formed by pairing an NCCT slice with its corresponding ground truth label, and the generated data, formed by pairing an NCCT slice with the artificial ground truth mask provided by the generator. The discriminator output is a probability that reflects the likelihood that the input sample is labeled with a real ground truth mask. The loss $L_D$ used to train the discriminator is defined as:

$$L_D = \mathrm{E}_{(x_L,y_L)\sim D_L}[\log D(y_L \oplus x_L)] + \mathrm{E}_{x_U \sim D_U}\big[\log\big(1 - D\big(S(x_U)\big) \oplus x_U\big)\big], \tag{7}$$

where $\oplus$ denotes concatenation along the channel dimension. With its ability to distinguish between real and artificial ground truth masks, the discriminator provides feedback that aids the generator to refine its output.

# III. Experimental Evaluation

This section details the publicly available dataset used, the experimental setup, the experimental results obtained, aimed to evaluate the actual contribution of each element of the proposed method to the segmentation accuracy obtained.

## A. Dataset

A publicly available dataset is used to experimentally evaluate the proposed method: the acute ischemic stroke dataset (AISD) [12]. AISD contains 397 NCCT scans acquired within 24 hours of acute ischemic stroke onset. The training set consists of 345 NCCT scans, whereas the remaining 52 scans are reserved for testing. The dataset provides segmentationnc masks for five infarct sub-classes: “remote infarct”, “clear acute infarct”, “blurred accute infarct”, “invisible acure infarct” and “infarct”. Except for the “invisible acute infarct” class, the remaining four classes can be collectively considered as a single class: “infarct lesion”. NCCT scans are provided in both .png and DICOM formats.

## B. Experimental Setup

For the semi-supervised setup, 30% of the training set is used as labeled data, while the remaining 70% is used as an unlabeled data pool. As the segmentation backbone, we utilize SCUNet++, a fully supervised architecture, originally tailored for pulmonary embolism, which integrates UNet++, multiple fusion dense skip connections, SWIN-Transformer attention mechanism and SWIN-UNet [13].

The discriminator is composed of four convolutional layers with kernel size 4×4 and channel dimensions of 64,

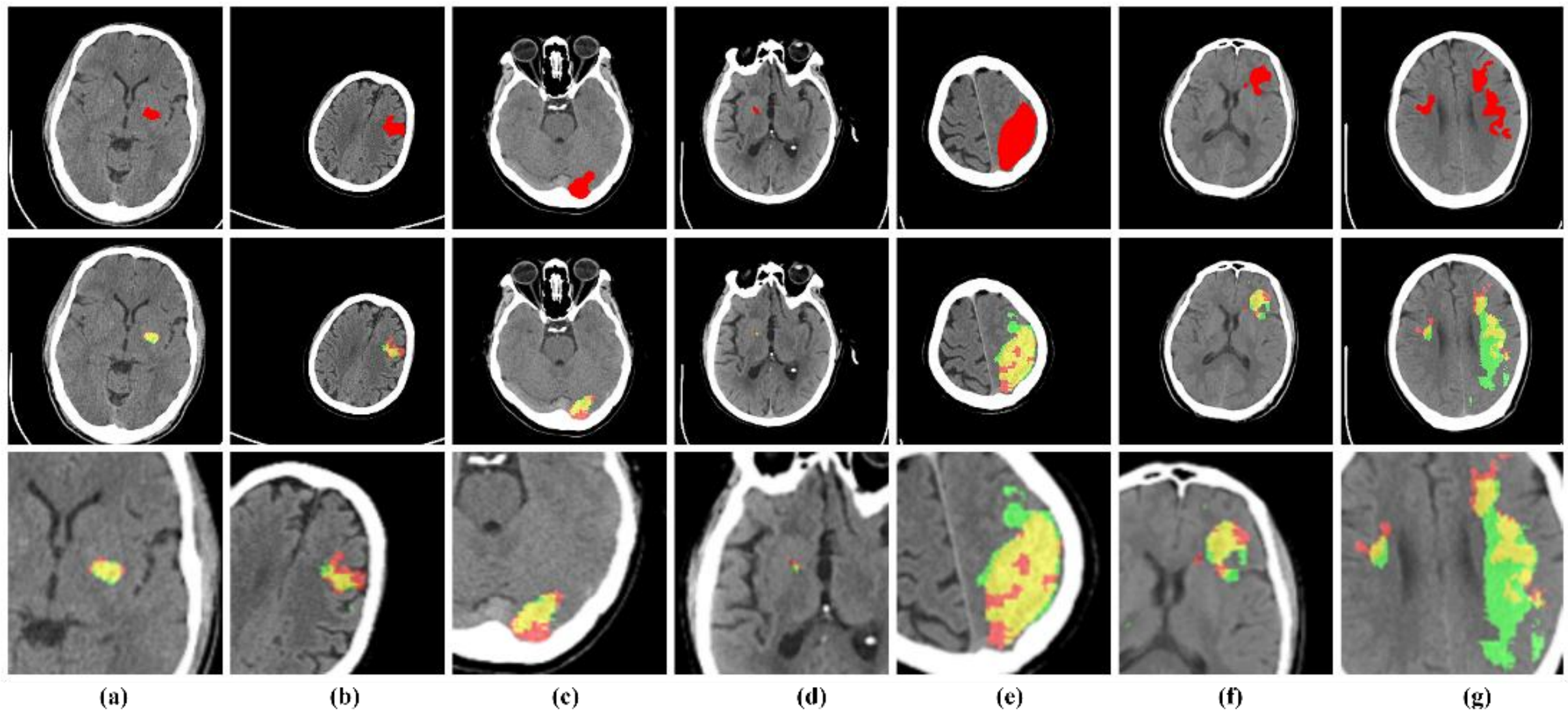

Fig. 2. Visualization of mask predictions at 0.3 labeled ratio. Top row: real ground truth masks of the test set. Middle row: Overlapped real ground truth and artificial ground truth masks (real ground truth pixels in red, artificial ground pixels is green, and overlap in yellow). Bottom row: Zoomed-in view of the overlapped masks.

128, 256, 512. Each convolutional layer is followed by a leaky ReLU, with a negative slope of 0.2, and a dropout layer with probability of 0.5, in order to mitigate overfitting and promote training stability. Moreover, the generator is trained using stochastic gradient descent (SGD), with a learning rate of $2.5 \times 10^{-4}$, whereas the discriminator is trained using Adam, with a learning rate of $10^{-4}$.

The weights of hybrid Dice loss, cross-entropy loss, feature matching loss and self-training loss are empirically determined as 0.6, 0.4, 0.1 and 1, respectively. The confidence threshold $\tau$ used to filter reliable predictions is set to 0.6. We use a batch size of 12 for all experiments. All networks were trained and evaluated on a workstation equipped with an NVIDIA GeForce RTX 3090 GPU.

### C. *Results*

In this section we quantitatively and qualitatively evaluate the results obtained by the proposed method, when applied to AISD. Our goal is to demonstrate the effectiveness in leveraging both labeled and unlabeled data.

Table I presents the results obtained under varying amounts of labeled data, using 10%, 30%, 50% and 80% of AISD labeled samples. Specifically, using only 30% of AISD labeled samples, the method achieves intersection over union (IoU) of 0.6585 and Dice score of 0.6797. As the amount of labeled data increases, both IoU and Dice scores increase, as expected. The best performance is observed when utilizing 80% of the labels of the training set with an IoU of 0.6788 and Dice score of 0.6975. It is noteworthy that the difference in Dice score between 30% and 80% labeled ratios is approximately 0.02, which is relatively small, when considering that in the latter case the method utilizes nearly all AISD labels.

TABLE I. EVALUATION AT DIFFERENT LABELED RATIOS

| Ratio | IoU | Dice | Recall | Precision |
|---|---|---|---|---|
| **0.1** | 0.5846 | 0.6107 | 0.6097 | 0.6561 |
| **0.3** | 0.6585 | 0.6797 | 0.6808 | 0.7200 |
| **0.5** | 0.6757 | 0.6965 | 0.6989 | 0.7261 |
| **0.8** | 0.6788 | 0.6975 | 0.7103 | 0.7218 |

TABLE II. EVALUATION AT DIFFERENT UNLABELED RATIOS (LABELED RATIO FIXED AT 30%)

| Ratio | IoU | Dice | Recall | Precision |
|---|---|---|---|---|
| **Fully-supervised** | 0.6339 | 0.6369 | 0.6341 | 0.6599 |
| **0.5** | 0.6277 | 0.6400 | 0.6325 | 0.6958 |
| **1** | 0.6585 | 0.6797 | 0.6808 | 0.7200 |

Aiming to evaluate the impact of unlabeled data in the proposed semi-supervised method, we conducted a series of experiments by varying the ratio of unlabeled samples used during training, whereas the labeled samples are fixed at 30% of available AISD labeled samples. This decision is supported by the results in Table I, where using the ground truth labels of 30% of AISD labeled data offers a reasonable trade-off between label efficiency and segmentation performance. In addition, we include a baseline experiment, in which only 30% of the labeled data are used in a fully supervised setting. This comparison allows us to assess the added value of incorporating unlabeled data during the training process. Table II presents the results for different ratios of unlabeled samples. Using half of the available unlabeled samples, the method achieves IoU of 0.6277 and a Dice score of 0.64, which is close to the fully supervised baseline, but still below the setting of using all the available unlabeled data. However, when all available unlabeled samples are used, the method reaches its best performance, resulting in IoU of 0.6585 and a Dice score of 0.6797. Notably, this setting outperforms the fully supervised model trained on the same labeled data.

Figure 2 provides a visualization of the proposed method's infarct lesion predictions. Yellow regions indicate the overlap between the ground truth (red) and the predicted mask (green), signifying areas of agreement between the model's output and the ground truth mask. It can be observed that the proposed method can handle a range of lesion sizes.

Examples of larger lesions are depicted in images (e) and (g). Specifically, for image (g), although the method's boundary delineation is not as precise as in image (e), the large lesion is correctly localized, whereas a smaller lesion in contralateral hemisphere is also identified. This highlights the model's capability for detecting multiple lesions within a single NCCT scan. Moreover, a very accurate localization and segmentation of a very small infarct lesion is illustrated in image (d). Furthermore, image (a), also contains a small lesion, demonstrating strong segmentation and boundary extraction. These visualizations affirm that, even under weak supervision, the model is capable of locating and segmenting lesions of various sizes with consistent performance.

## IV. CONCLUSIONS

In this work, we present the application of a semi-supervised, GAN-based framework to the challenging task of segmenting ischemic lesions in early-phase NCCT scans, addressing the critical need for sensitive diagnostic tools for hyperacute stroke, where ischemic indicators on NCCT are often minimal. The main strength of the proposed segmentation method lies in its semi-supervised design, which mitigates the dependency on extensive, pixel-level annotations. The proposed method effectively utilizes both labeled and unlabeled data. The integration of a hybrid loss function and adversarial training aids accurate delineation, generating anatomically plausible results.

The performance of the proposed method on the AISD dataset confirms its potential as a powerful assistive tool in the stroke diagnosis workflow, which enhances the detection of early-stage ischemia and facilitates rapid and reliable clinical decision-making. The experimental evaluation leads to the following conclusions:

*i.* A fully-supervised network, like SCUNet++, can achieve comparable, or even superior results in the NCCT image segmentation task, when integrated within a semi-supervised GAN-based framework. As shown in Section III.C, even with only 30% of the total labeled samples in the AISD dataset, the proposed method accurately localizes and delineates infarct lesions, surpassing the segmentation accuracy of the fully-supervised network.

*ii.* In the hyperacute phase, the signs of ischemia, such as cytotoxic edema causing subtle hypoattenuation, can be extremely faint and difficult to detect. The high IoU and Dice scores achieved by the proposed method demonstrate its strong capability to accurately extract even these challenging infarct lesions, as illustrated in Fig 2.

*iii.* The proposed method effectively leverages not only the supervised signal from labeled samples but also crucial information from unlabeled samples, and accurately localizes and delineates infarct lesions, regardless of their number or size. Ultimately, the proposed method relies less on costly and labor-intensive annotations, while it maintains high segmentation accuracy.

Future work will focus on validating the proposed method across more diverse NCCT datasets and investigating its integration into clinical workflows. In the long term, this research supports the development of AI-powered solutions that streamline the stroke care pathway and enhance patient outcomes.

## ACKNOWLEDGMENT

This work was partially funded by the European project SEARCH (https://ihi-search.eu/), which is supported by the Innovative Health Initiative Joint Undertaking (IHI JU) under grant agreement No. 101172997. The JU receives support from the European Union's Horizon Europe research and innovation programme and COCIR, EFPIA, Europa Bio, MedTech Europe, Vaccines Europe, Medical Values GmbH, Corsano Health BV, Syntheticus AG, Maggioli SpA, Motilent Ltd, Ubitech Ltd, Hemex Benelux, Hellenic Healthcare Group, German Oncology Center, Byte Solutions Unlimited, AdaptIT GmbH. Views and opinions expressed are however those of the author(s) only and do not necessarily reflect those of the aforementioned parties. Neither of the aforementioned parties can be held responsible for them.